\documentclass{article}
\PassOptionsToPackage{numbers, compress}{natbib}
\usepackage[preprint]{neurips_2023}

\usepackage[T1]{fontenc}

\usepackage{amsmath,amssymb,amsfonts}
\usepackage{algorithmic}
\usepackage{graphicx}
\usepackage{textcomp}
\usepackage{xcolor}
\usepackage{hyperref}       
\usepackage{url}            
\usepackage{amsfonts}       
\usepackage{nicefrac}       
\usepackage{graphicx}
\usepackage{caption}
\usepackage{subcaption}
\usepackage{multirow}
\usepackage[ruled,noend,linesnumbered]{algorithm2e}
\usepackage{amssymb}
\usepackage{booktabs}
\usepackage[dvipsnames]{xcolor}
\usepackage{amsmath}

\usepackage{latexsym}
\usepackage{amssymb}
\usepackage{amsmath}
\usepackage{amsthm}
\usepackage{booktabs}
\usepackage{enumitem}
\usepackage{graphicx}
\usepackage{color}
\usepackage[ruled,linesnumbered]{algorithm2e}
\usepackage{adjustbox}

\usepackage[dvipsnames]{xcolor}
\usepackage{url}

\usepackage{tikz}
\usetikzlibrary{arrows.meta}
\usetikzlibrary{shapes.geometric,shapes.misc,arrows.meta}

\tikzset{%
    flows/.style={dotted,red!90!black,ultra thick},
    open_flows/.style={dashed,blue!80!black,ultra thick},
    back_prec/.style={gray!80},
    back_flows/.style={dotted,red!90!black!50!white,ultra thick},
    resource/.style={green!70!black, thick,dashdotted},
    back_resource/.style={green!90!black, thick,dashdotted},
    pooling/.style={cyan, thick,dashdotted},
    back_pooling/.style={cyan!70!blue, thick,dashdotted},
    self_loop_task/.style={red},
    self_loop_resource/.style={ForestGreen},
    self_loop_pool/.style={cyan!80!black}
}

\newtheorem{example}{Example}

\usepackage{todonotes}

\newcommand{\mc}[1]{\mathcal{#1}}

\newcommand{\mi}[1]{\mathit{#1}}
\newcommand{\tasks}{\mc{T}}
\newcommand{\resources}{\mc{R}}
\newcommand{\precedences}{\mc{P}}

\begin{document}

\title{Learning to Solve Resource-Constrained Project Scheduling Problems with Duration Uncertainty using Graph Neural Networks
\thanks{Our work has benefitted from the AI Interdisciplinary Institute ANITI funded by the French "Investing for the Future – PIA3" program under the Grant agreement n°ANR-23-IACL-0002.}
}

\author{%
  Guillaume Infantes\\
  Jolibrain, Toulouse, France\\
  \texttt{guillaume.infantes@jolibrain.com}
  \And
  Stéphanie Roussel\\
  DTIS/ONERA, Université de Toulouse, France\\
  \texttt{stephanie.roussel@onera.fr}\\
  \And
  Antoine Jacquet\\
  Jolibrain, Toulouse, France\\
  \texttt{antoine.jacquet@jolibrain.com}
  \And
  Emmanuel Benazera\\
  Jolibrain, Toulouse, France\\
  \texttt{emmanuel.benazera@jolibrain.com}
}
\maketitle

\maketitle

\begin{abstract}
The Resource-Constrained Project Scheduling Problem (RCPSP) is a classical scheduling problem that has received significant attention due to of its numerous applications in industry. However, in practice, task durations are subject to uncertainty that must be considered in order to propose resilient scheduling. In this paper, we address the RCPSP variant with uncertain tasks duration (modeled using known probabilities) and aim to minimize the overall expected project duration. Our objective is to produce a baseline schedule that can be reused multiple times in an industrial setting regardless of the actual duration scenario. We leverage Graph Neural Networks in conjunction with Deep Reinforcement Learning (DRL) to develop an effective policy for task scheduling. This policy operates similarly to a priority dispatch rule and is paired with a Serial Schedule Generation Scheme to produce a schedule. Our empirical evaluation on standard benchmarks demonstrates the approach's superiority in terms of performance and its ability to generalize. The developed framework, Wheatley, is made publicly available online to facilitate further research and reproducibility.
\end{abstract}

\section{Introduction}

Project scheduling is a critical component in diverse industries, including aerospace, construction and manufacturing. In this context, the Resource-Constrained Project Scheduling Problem (RCPSP) serves as a classical paradigm, aiming at optimizing tasks schedules under limited resources constraints. 
The RCPSP has been extensively explored in the literature, encompassing a wide range of solving techniques, ranging from exact methods to metaheuristics, as well as studies of its numerous variants, as detailed in \cite{ARTIGUES2025} and \cite{HARTMANN20221}.

In practice, task durations are often subject to uncertainty coming from several events (equipment failures, supply chain delays, faults occurring during the tasks execution, etc.). To mitigate this, two main paradigms are commonly employed: \emph{reactive} and \emph{proactive} approaches. Reactive scheduling offers flexibility through dynamic responses to real-time deviations, but may lead to unstable schedules with significant variability, hindering repeatability. In contrast, proactive scheduling establishes a fixed schedule in advance (offline scheduling), designed to accommodate a range of possible outcomes while limiting online adaptation. This characteristic makes proactive scheduling particularly valuable in settings where a plan must be applied repeatedly across multiple scenarios. A typical example arises in the manufacturing sector, where a standardized process is consistently repeated across each production cycle, yet the actual task durations can fluctuate due to inherent operational uncertainties.

In order to accomodate delays in the baseline schedule produced by proactive approaches, Serial Schedule Generation Schemes (SGSs) can be employed in combination with Priority Dispatch Rules (PDRs) \cite{kolisch1996serial}. In a SGS, tasks are scheduled one at a time in a predefined sequence, ensuring that resource and precedence constraints are satisfied at each step. The task ordering is determined by a priority rule, which typically assigns rankings based on tasks and resources attributes. Building upon recent advances in Reinforcement Learning (RL) and Graph Neural Network (GNN), this work seeks to learn a priority dispatch rule in the form of a policy. This policy will determine the next task to append to the priority list, given a set of candidate tasks. 

Our paper presents the following key contributions:
\begin{itemize}
    \item We model the RCPSP with uncertainty as a Markov Decision Process (MDP) so that the sequence of states to reach a final state is mapped to a priority rule; 
    \item We define a GNN-based agent aligned with the MDP formulation and provide in-depth representation details;
    \item We evaluate our approach against baseline methods, including both Priority Dispatch Rules (PDRs) and deterministic approaches, demonstrating its performance and generalizability to unseen datasets;
    \item We provide the open-source framework Wheatley \cite{wheatley} for future research and  reproducibility.
\end{itemize}

This paper is organized as follows. Section~\ref{sec:related} provides an overview of related works. Section~\ref{sec:mdp} presents the MDP formulation of the problem, followed by Section~\ref{sec:gnn} which focuses on the GNN-based approach. The experimental setup and results are outlined in Section~\ref{sec:expes}. Finally, we conclude our findings and discuss future perspectives in Section~\ref{sec:conclusion}.

\section{Related Works}
\label{sec:related}

Because of its numerous applications in industry, RCPSP has received significant attention over the past fifty years. 
Recent surveys, notably \cite{ARTIGUES2025}, highlights the range of solving techniques that have been developed for solving this problem. Additionally, surveys like \cite{HARTMANN20221} provide an overview of the main variants explored. This section concentrates on three key aspects of our problem and approach: priority dispatch rules design combined with SGS, considering duration uncertainty within RCPSP and leveraging Machine Learning (ML) for scheduling problem resolution. 

Combining PDRs within SGSs has been studied for decades \cite{kolisch1996serial}. Recently, ML techniques have been employed to select the optimal priority rule. In \cite{GUO2021114116}, a decision tree was developed in order to choose the best PDR for each instance. The best PDR can also be selected at each step of the scheduling, as demonstrated by a multilayer feed-forward neural network in \cite{golab:hal-04163409} or through RL in  \cite{WANG20222144}. Other studies, such as \cite{LUO2022116753, djumic2021ensembles, regnier2021empirical} have proposed approaches to design new PDRs, typically using Genetic Programming (GP). 

When accounting for uncertainty, \cite{ZHANG2025106816} presents a GP hyper-heuristic that automatically evolves priority rules for the RCPSP with stochastic durations and transfer times. In \cite{GOLI2024109427}, multiple uncertainty parameters (duration, risk and quality of each activity) are considered within GP to tackle a multi-objective variant of the RCPSP. \cite{TeichteilKonigsbuch23} considers the RCPSP with uncertain task duration using GNN to capture the problem's structure. The proposed approach, SIRENE, leverages an exact Constraint Programming solver and learn to predict the activity start dates, while resolving inconsistencies via a SGS. Notably, their method operates within a reactive context, whereas our focus lies in developing a proactive scheduling policy. 

Leveraging RL along with GNNs in the context of scheduling is not novel. For example, several studies have addressed the Job Shop Scheduling Problem (JSSP) and its variants, including dynamic \cite{liang2023dynamic} and uncertain cases \cite{InfantesRPJB24}. 
However, adapting those approaches to RCPSPs is not trivial. Unlike JSSPs, where priority rules are equivalent to precedences, RCPSPs require the GNN to account for the resources capacities constraints. 
In the context of RCPSPs, \cite{VerhaegheCPQ24} employs GNNs to learn flow precedences in the graph. The work by \cite{Zhao2022} shares similarities our approach, as it models a RCPSP using a MDP to develop a GNN-based agent for learning optimal priority rules. Nevertheless, there are key differences. First, uncertainty is not considered in their work, which affects the MDP and RL approach. Second, our modeling differs significantly, particularly in the state and reward design, as well as GNN problem representation. Finally, to the best of our knowledge, their approach is not publicly available, hindering  reproducibility and the comparative analysis with their work.

	\section{RCPSP with uncertainty as a MDP}
        \label{sec:mdp}
	
	We extend an Activity-on-Node-flow (AON-flow) formulation of the static RCPSP problem, as initially described in \cite{ARTIGUES2003249}. We first outline the RCPSP model, followed by the AON-flow formulation, then and the associated MDP. 
	
	\subsection{Background}
	
	The RCPSP can be modeled by a tuple $(\tasks, \resources, \precedences)$, in which a set of tasks $\tasks$ has to be scheduled on a set of resources set $\resources$ while respecting precedence relationship $\precedences$. Each resource $r \in \resources$ has a capacity $\mi{capa}_r$ and is renewable. Each task $t \in \tasks$ has a non-negative duration $\delta_t$, and for each resource $r \in \resources$, a non-negative consumption $\mi{cons}_{t,r}$ indicates the amount of resource $r$ utilized by task $t$. The precedence relationship $\precedences$ is a set of tasks pair $(t_1,t_2) \in \tasks^2$ representing that  task $t_1$ must be completed before $t_2$ can start. 
	
	We suppose that $\tasks$ contains two specific tasks $\alpha$ and $\omega$ that represent the start and the end of the project. Formally, $\alpha$ and $\omega$ have a null duration, consume resources to their whole capacity and respectively precede and follow all other tasks. 
	
	A function $\sigma$ that assigns a start date $\sigma(t)$ to each task $t \in \tasks$ (also called a schedule) is a solution if it is precedence-feasible and resource-feasible. An optimal solution of the RCPSP is a solution that minimizes the makespan, \textit{i.e.} the date of task $\omega$.  
	
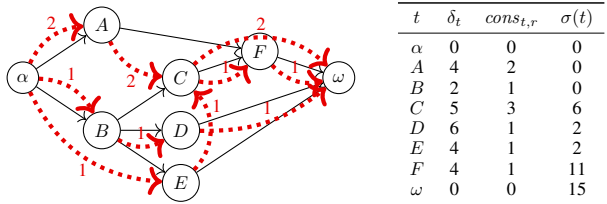
\begin{figure}[h]
			\begin{tikzpicture}[scale=0.7, every node/.style={scale=0.7}]
				\node[circle,draw] (alpha) at (0,0) {$\alpha$};
				\node[circle,draw] (a) at (1.5,1) {$A$};
				\node[circle,draw] (b) at (1.5,-1) {$B$};
				\node[circle,draw] (c) at (3,0) {$C$};
				\node[circle,draw] (d) at (3,-1) {$D$};
				\node[circle,draw] (e) at (3,-2) {$E$};
				\node[circle,draw] (f) at (4.5,.5) {$F$};
				\node[circle,draw] (omega) at (6,0) {$\omega$};
				\draw[->] (alpha) -- (a);
				\draw[->] (alpha) -- (b);
				\draw[->] (a) -- (f);
				\draw[->] (b) -- (c);
				\draw[->] (b) -- (d);
				\draw[->] (b) -- (e);
				\draw[->] (c) -- (f);
				\draw[->] (f) -- (omega);
				\draw[->] (d) -- (omega);
				\draw[->] (e) -- (omega);

                \path[->, dotted,red!90!black,ultra thick] (alpha) edge[bend left] node[above] {2} (a);
                    \path[->, dotted,red!90!black,ultra thick] (alpha) edge[bend left] node[above] {1} (b);
                    \path[->, dotted,red!90!black,ultra thick] (alpha) edge[bend right] node[below] {1} (e);
                    \path[->, dotted,red!90!black,ultra thick] (b) edge[bend right] node[above] {1} (d);
                    \path[->, dotted,red!90!black,ultra thick] (e) edge[bend right=45] node[right,pos=.8] {1} (c);
                    \path[->, dotted,red!90!black,ultra thick] (a) edge[bend right] node[below] {2} (c);
                    \path[->, dotted,red!90!black,ultra thick] (c) edge[bend right] node[above] {1} (f);
                    \path[->, dotted,red!90!black,ultra thick] (d) edge[bend right=20] node[above] {1} (omega);
                    \path[->, dotted,red!90!black,ultra thick] (f) edge[bend right] node[above] {1} (omega);
                    \path[->, dotted,red!90!black,ultra thick] (c) edge[bend left=45] node[above] {2} (omega);
				
				\node[anchor=west] at (7,-.5) {
						\begin{tabular}{cccc}
						\toprule
						$t$ & $\delta_t$ & $\mi{cons}_{t,r}$ & $\sigma(t)$ \\
						\midrule
						$\alpha$ & 0 & 0 & 0 \\
						$A$ & 4 & 2 & 0 \\ 
						$B$ & 2 & 1 & 0 \\
						$C$ & 5 & 3 & 6 \\
						$D$ & 6 & 1 & 2 \\
						$E$ & 4 & 1 & 2 \\
						$F$ & 4 & 1 & 11 \\
						$\omega$ & 0 & 0 & 15\\
						\bottomrule
					\end{tabular}
				};
			\end{tikzpicture}
			\caption{Example tasks features (precedences are black arrows of the graph, duration and resource consumption in the table) and schedule $\sigma$ in the table. Flow arcs mentioned in Example~\ref{ex:flows} are depicted in dotted red.}
			\label{fig:precedence}
	\end{figure}

	\begin{example} \label{ex:description}
		Based on the example presented in \cite{ARTIGUES2003249}, we consider a toy example with 8 tasks $\tasks = \{\alpha,A,B,C,D,E,F,\omega\}$ and whose features are described in Figure~\ref{fig:precedence}. We consider a unique resource $r$ with a capacity $\mi{capa}_r=4$. A schedule is given in the figure, with an associated makespan equal to 15.
	\end{example}
	
	The AON formulation of a RCPSP instance is a directed acyclic graph in which there is a node for each task $t \in \tasks$ and an arc for each precedence $(t_1,t_2) \in \precedences$. Such a graph can be extended into an AON-flow graph, in order to represent schedules \cite{ARTIGUES2003249}. Formally, the AON-flow graph contains additional arcs, here called flow arcs. A flow arc is a tuple $(r,k,t_1,t_2)$ that indicates that $k$ units of resource $r$ are transferred from task $t_1$ to task $t_2$. Intuitively, if a task $t$ requires $\mi{cons}_{t,r} > 0$ units of resource $r$ to be executed, then the AON-flow graph contains $l$ flow arcs $(r,k,t_1,t)$ entering task $t$ such that the sum of the units transferred $k$ over those $l$ arcs is equal to $\mi{cons}_{t,r}$, and also contains $l'$ flow arcs $(r,k,t,t_2)$ out-going task $t$ such that the sum of transferred units over those $l'$ arcs is also equal to $\mi{cons}_{t,r}$. 
	
	The AON-flow graph intuitively represents a partial order for the task execution, and for which two solutions can be easily generated (earliest starting dates solution when left-shifting and latest ending dates when right-shifting) (see \cite{ARTIGUES2003249} for more details on that step). 

    \begin{example} \label{ex:flows}
        The AON-flow graph associated with Example~\ref{ex:description} is illustrated on Figure~\ref{fig:precedence} with dotted red arrows. Flow arcs are represented through red dotted arrows. Note that there might exist several AON-flow graphs associated with the schedule of Figure~\ref{fig:precedence}, but there are all equivalent with respect to their makespan values. 
    \end{example}

	\subsection{MDP Formulation of Deterministic RCPSP}
	
	In this work, we utilize the AON-flow graph to represent a partial schedule where tasks start as early as possible. Our objective is to iteratively build a complete schedule, starting from a flow-free graph (initial state), and progressively selecting and inserting tasks into the schedule by adding associated flows (action and transition), until all tasks are scheduled (final state). This process is formally captured through the following deterministic MDP.   
	
	\textbf{States.}  A state $s_i$ is a specific AON-flow graph represented by the tuple $(\tasks,$ $\mi{Sched}_i,$ $\sigma_i,$ $\precedences,$  $(\mi{Flow}_{r,i})_{r \in \mc{R}}, (\mi{Open}_{r,i})_{r \in \mc{R}})$. $\tasks$ is the set of nodes, which is in one-to-one correspondence with the set of project tasks. $\mi{Sched}_i$ is the set of tasks that are already scheduled in state $s_i$. A start date is assigned to each task in $s_i$ via the function $\sigma_i$. We consider three types of arcs. The precedence relationships between tasks are embodied by $\precedences$, denoting the first set of tasks. For each resource $r$, $\mi{Flow}_{r,i}$  represents a second set of arcs, specifically the flows arcs of $r$, where each flow arc $(k,t_1,t_2)$ signifies that $k$ units of resource $r$ are transferred from task $t_1$ to task $t_2$. Lastly, $\mi{Open}_{r,i}$ denotes the open flow arcs of resource $r$ at step $i$ (they have an origin but no destination), representing flows released by scheduled tasks but not yet fully consumed by other tasks. Formally, an open flow arc in $\mi{Open}_{r,i}$ is a pair $(t,k)$ indicating that $k$ units of resource $r$ that task $t$ await consumption.
	
	The initial state $s_0$ is formally defined by: $\mi{Sched}_0 = \{\alpha\}$; $\sigma(\alpha)=0$; for all $r\in\mc{R}$, $\mi{Flow}_{r,0} = \emptyset$; for all $r\in\mc{R}$, $\mi{Open}_{r,0} = \{(\alpha,\mi{capa}_r)\}$, that represents the fact that all resources units are initially available.
	The terminal state is reached at step $|\mc{T}|$, when all tasks are scheduled. 
	
	\textbf{Actions. } An action consists in selecting a single task to be inserted in the schedule. The set of available actions $A_i$ at state $s_i$ comprises tasks whose predecessors have all been scheduled. Formally, $A_i = \{t \in \tasks | \precedences_t \subseteq \mi{Sched}_i \}\}$, where $\precedences_t = \{t_{p} \in \tasks | (t_p, t) \in \precedences\}$, i.e. all tasks preceding $t$. 
	
	\textbf{Transitions. } The transition function consists in updating the partial schedule of state $s_i$ by inserting a task $t \in A_i$ selected through action $a_i$. Such a transition can be seen as an addition of a task as done in a Serial SGS (SSGS). We detail that process here, as it supports the precise implementation of the problem, as detailed later. 
	
Intuitively, the resulting state $s_{i+1}$ consists in the insertion of task $t$ as early as possible, with the earliest open flows, and update flows and open flows accordingly. The mechanism to compute the next state is detailed in Algorithm~\ref{alg:updateState}. At line~\ref{line:prev_set}, $t$ is added to the set of scheduled tasks. We also initialize a set $\mi{Prev}_t$ that will contain all tasks preceding $t$, either through precedences $\precedences$ or by flow precedences. This set is set up with $\precedences_t$ (tasks preceding $t$ according to $\precedences$). Then, for each resource $r$ consumed by $t$, we first order the set of open flows $\mi{Open}_{r,i}$ by their availability date, \textit{i.e.} flow $(t_1,k_1)$ is before flow $(t_2,k_2)$ in the ordered list if $t_1$ ends before $t_2$ ($\sigma_i (t_1) < \sigma_i (t_2)$). Tie breaks is done using a lexicographic order over tasks. Then, lines~\ref{line:init_flow}-\ref{line:update_cons} allow to compute the earliest list $F_{i,r}$ of open flows that allow to reach $\mi{cons_{t,r}}$. All tasks involved in $F_{i,r}$ are flow predecessors of $t$ and added to $\mi{Prev}_t$ (line~\ref{line:update_prev}). The new set of open flows $\mi{Open}_{r,i+1}$ is the set $\mi{Open}_{r,i}$ from which all flows of $F_{i,r}$ are removed and that contains a new open flow for task $t$ and resource $r$. Lines~\ref{line:consGreater}-\ref{line:update_open_rem_flow} allow to handle the case where flows of $F_{i,r}$ do not match exactly the required consumption $\mi{cons}_{t,r}$. In that case, the last flow of $F_{i,r}$ is split into two flows: the first one contains exactly what is needed for $t$ and remains in $F_{i,r}$ (line~\ref{line:new_flow_in_fir}), the second contains the remaining units of $r$ and is added to $\mi{Open}_{r,i+1}$ (line~\ref{line:update_open_rem_flow}). The new set of flows $\mi{Flow_{r,i+1}}$ is equal to the set $\mi{Flow_{r,i+1}}$ to which are added all flows in $F_{i,r}$ (line~\ref{line:update_flow_set}). Finally, task $t$ starts at the latest date of all its predecessors (line~\ref{line:compute_date}), which is updated in $\sigma_{i+1}$ (line~\ref{line:update_sigma}).

\begin{algorithm}
	\caption{Compute Next State}\label{alg:updateState}
	\KwData{State $s_i$, task to insert $t$}
	\KwResult{State $s_{i+1}$}
	$\mi{Sched}_{i+1} \gets \mi{Sched}_i \cup \{t\}$ ;
	$\mi{Prev}_t \gets \precedences_t$ \; \label{line:prev_set}
	\ForEach{$r \in \resources$ s.t. $\mi{cons}_{t,r} > 0$}{
		$\mi{OrderedOpen}_{r,i} \gets \mathsf{OrderEarliest}(\mi{Open}_{r,i})$\; \label{line:order_open}
		$F_{i,r} \gets []$, $\mi{cons} \gets 0$ \; \label{line:init_flow}
		\While{$\mi{cons} < \mi{cons}_{t,r}$}{ \label{line:start_while}
			$(t_p,k_p) \gets \mathsf{RemoveFirst}(\mi{OrderedOpen}_{r,i})$ \; \label{line:getfirstflow}
			$\mi{Prev}_t \gets \mi{Prev}_t \cup \{t_p\}$ \; \label{line:update_prev}
			$F_{i,r} \gets \mathsf{Append}(F_{i,r}, (t_p, k_p))$ \; \label{line:append_fir}
			$\mi{cons} \gets \mi{cons} + k_p$ \; \label{line:update_cons}
		}
		$\mi{Open}_{r,i+1} \gets \mi{Open}_{r,i} \setminus F_{i,r} \cup \{(t, \mi{cons}_{t,r})\}$ \; \label{line:update_open}
		\If{$\mi{cons} > \mi{cons}_{t,r}$}{ \label{line:consGreater}
			$k_\mi{rem} \gets \mi{cons} - \mi{cons}_{t,r}$ \; \label{line:consRemain}
			$(t_l,k_l) \gets \mathsf{RemoveLast}(F_{i,r})$ \; \label{line:lastFlow}
			$F_{i,r} \gets \mathsf{Append}(F_{i,r}, (t_l, k_l - k_\mi{rem}))$ \; \label{line:new_flow_in_fir}
			$\mi{Open}_{r,i+1} \gets \mi{Open}_{r,i+1} \cup \{(t_l, k_\mi{rem})\}$ \label{line:update_open_rem_flow}
		}
		$\mi{Flow}_{i+1,r} \gets \mi{Flow}_{i,r} \cup F_{i,r}$ \; \label{line:update_flow_set}
	}
	$\mi{date}_t \gets \mi{max}_{t_p \in \mi{Prev}_t} (\sigma_i(t_p) + \delta_ {t_p})$\; \label{line:compute_date}
	$\sigma_{i+1} \gets \sigma_i \cup \{(t,\mi{date}_t)\}$ \label{line:update_sigma}
\end{algorithm}
	
	\textbf{Reward} The reward is equal to $0$ for all states except for the terminal state where the reward is equal to the opposite of the makespan, \textit{i.e.} $-\sigma_{|\mc{T}|}(\omega)$. In the following, we write $\sigma$ instead of $\sigma_{|\mc{T}|}$ as all tasks have an associated start date. 

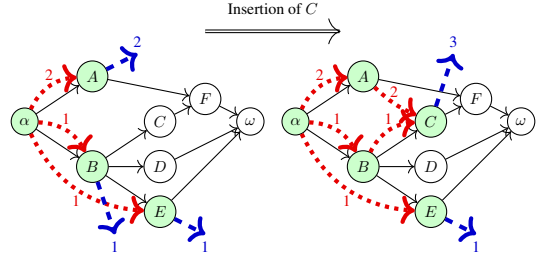
\begin{figure}
        \centering
        \begin{tikzpicture}[scale=0.6, every node/.style={scale=0.6}]
            \begin{scope}
				\node[circle,draw,fill=green!20] (alpha) at (0,0) {$\alpha$};
				\node[circle,draw,fill=green!20] (a) at (1.5,1) {$A$};
				\node[circle,draw,fill=green!20] (b) at (1.5,-1) {$B$};
				\node[circle,draw] (c) at (3,0) {$C$};
				\node[circle,draw] (d) at (3,-1) {$D$};
				\node[circle,draw,fill=green!20] (e) at (3,-2) {$E$};
				\node[circle,draw] (f) at (4,.5) {$F$};
				\node[circle,draw] (omega) at (5,0) {$\omega$};
				\draw[->] (alpha) -- (a);
				\draw[->] (alpha) -- (b);
				\draw[->] (a) -- (f);
				\draw[->] (b) -- (c);
				\draw[->] (b) -- (d);
				\draw[->] (b) -- (e);
				\draw[->] (c) -- (f);
				\draw[->] (f) -- (omega);
				\draw[->] (d) -- (omega);
				\draw[->] (e) -- (omega);

                    \path[->, flows] (alpha) edge[bend left] node[above] {2} (a);
                    \path[->, flows] (alpha) edge[bend left] node[above] {1} (b);
                    \path[->, flows] (alpha) edge[bend right] node[below] {1} (e);
                    \path[->, dashed,blue!80!black,ultra thick] (b) edge[] node[below,pos=1] {1} (2,-2.5);
                    \path[->, dashed,blue!80!black,ultra thick] (e) edge[] node[pos=1,below] {1} (4,-2.5);
                    \path[->, dashed,blue!80!black,ultra thick] (a) edge[] node[above,pos=1] {2} (2.5,1.5);
                \end{scope}

                \begin{scope}[xshift=6cm]
				\node[circle,draw,fill=green!20] (alpha) at (0,0) {$\alpha$};
				\node[circle,draw,fill=green!20] (a) at (1.5,1) {$A$};
				\node[circle,draw,fill=green!20] (b) at (1.5,-1) {$B$};
				\node[circle,draw,fill=green!20] (c) at (3,0) {$C$};
				\node[circle,draw] (d) at (3,-1) {$D$};
				\node[circle,draw,fill=green!20] (e) at (3,-2) {$E$};
				\node[circle,draw] (f) at (4,.5) {$F$};
				\node[circle,draw] (omega) at (5,0) {$\omega$};
				\draw[->] (alpha) -- (a);
				\draw[->] (alpha) -- (b);
				\draw[->] (a) -- (f);
				\draw[->] (b) -- (c);
				\draw[->] (b) -- (d);
				\draw[->] (b) -- (e);
				\draw[->] (c) -- (f);
				\draw[->] (f) -- (omega);
				\draw[->] (d) -- (omega);
				\draw[->] (e) -- (omega);

                    \path[->, dotted,red!90!black,ultra thick] (alpha) edge[bend left] node[above] {2} (a);
                    \path[->, dotted,red!90!black,ultra thick] (alpha) edge[bend left] node[above] {1} (b);
                    \path[->, dotted,red!90!black,ultra thick] (alpha) edge[bend right] node[below] {1} (e);
                     \path[->, dotted,red!90!black,ultra thick] (b) edge[bend left] node[above] {1} (c);
                    \path[->, dotted,red!90!black,ultra thick] (a) edge[bend right=10] node[above] {2} (c);
                    \path[->, dashed,blue!80!black,ultra thick] (e) edge[] node[pos=1,below] {1} (4,-2.5);
                    \path[->, dashed,blue!80!black,ultra thick] (c) edge[] node[above,pos=1] {3} (3.5,1.5);
                \end{scope}

                \draw[double,->] (4,2) -- (7,2);
                \node at (5.5,2.5) {Insertion of $C$};
                
			\end{tikzpicture}
        \caption{Scheduled tasks (light green), flows (dotted red) and open flows (dashed blue) for states $s_i$ and $s_{i+1}$ described in Example~\ref{ex:transition} } 
        \label{fig:AON-flow_insertion}
    \end{figure}

    \begin{example} \label{ex:transition}
        Following Example~\ref{ex:description}, $s_i$ is a state where tasks $\alpha$, $A$, $B$ and $E$ have been scheduled, with respective start dates $0$, $0$, $0$ and $2$. The associated flows and open flows are respectively represented in dotted red and dashed blue arcs of the left part of Figure~\ref{fig:AON-flow_insertion}. The candidate actions set $A_i$ is composed of tasks $C$ and $D$, \textit{i.e.} tasks for which all predecessors are scheduled and not scheduled yet. If task $C$ is selected for insertion, then the resulting flows in state $s_{i+1}$ are on the right of Figure~\ref{fig:AON-flow_insertion}: the earliest open flows in $s_i$ reaching $C$'s consumption are $(A,2)$ and $(B,1)$. Those become flows of state $s_{i+1}$ and the open flow $(C,3)$ is added.
    \end{example}

	\subsection{Representing Uncertainty}
       \label{sec:repr_unc}
	In this paper, we consider uncertainty in task duration. For each task $t \in \tasks$, the duration $\delta_t$ is replaced by a random variable $\Delta_t$ following a probability distribution $\mathbb{D}_t$. For each task $t$, we suppose that the agent knows three elements of distribution $\mathbb{D}_t$: the minimum duration of task $t$, its maximum duration and its mode duration, respectively denoted $\delta_t^\mi{min}$, $\delta_t^\mi{max}$ and $\delta_t^\mi{mod}$. The mode duration $\delta_t^\mi{mod}$ represents the duration with the highest probability, with tiebreaks favoring the duration closest to the expected duration. The resulting objective is to minimize the expected makespan, represented by the expected completion time of $\omega$.
	
	To account for this uncertainty, we extend the previously defined MDP by introducing three distinct graph flows, each corresponding to a specific duration type within the set $\Delta = \{\mi{min}, \mi{max}, \mi{mod}\}$. Formally, the MDP contents are updated as follows:
	\begin{itemize}
		\item a state becomes a tuple $(\tasks,$ $ \mi{Sched}_i,$ $(\sigma_i^d)_{d \in \Delta},$ $\precedences,$  $(\mi{Flow}_{r,i}^d)_{r \in \mc{R}, d \in \Delta},$ $(\mi{Open}_{r,i}^d)_{r \in \mc{R}, d \in \Delta})$ that contains a schedule $(\sigma_i^d)$, open flows $(\mi{Open}_{r,i}^d)$ and complete flows $(\mi{Flow}_{r,i}^d)$ for each duration type;
		\item each flow graph associated with a duration type $d$ is updated independently upon a task insertion, following Algorithm~\ref{alg:updateState};
		\item the reward of the terminal state is computed by sampling a duration $\tilde d \sim \mathbb{D}_t$. Starting with the initial state, the sequence of actions decided by the agent is applied until the terminal state is reached, yielding a graph flow and a schedule $\tilde \sigma$ with sampled duration values. The reward provided to the agent is the opposite of the makespan associated with $\tilde \sigma$, i.e. $-\tilde \sigma(\omega)$.  
	\end{itemize}
	
	
	\section{GNN-based Agent}
        \label{sec:gnn}
       In this section, we provide an in-depth view of the agent that computes the policy based on current partial schedule, observed as a flow graph. One pivotal aspect of this agent is its capacity to effectively handle graphs, here processed with a GNN.

       \subsection{Overview}
       In order to use RL algorithms like PPO \cite{PPO}, our agent must output, at each step, a probability distribution over candidate actions (stochastic policy), and an estimate of the expected future outcomes (value). For the action probability, we leverage the output logits from the GNN nodes. To compute probabilities only on candidates actions (tasks whose predecessors are already scheduled), we employ action masking \cite{ppo-mask}. 
       In contrast, computing the value necessitates a holistic graph assessment, which  requires a global graph pooling operator, defined below.

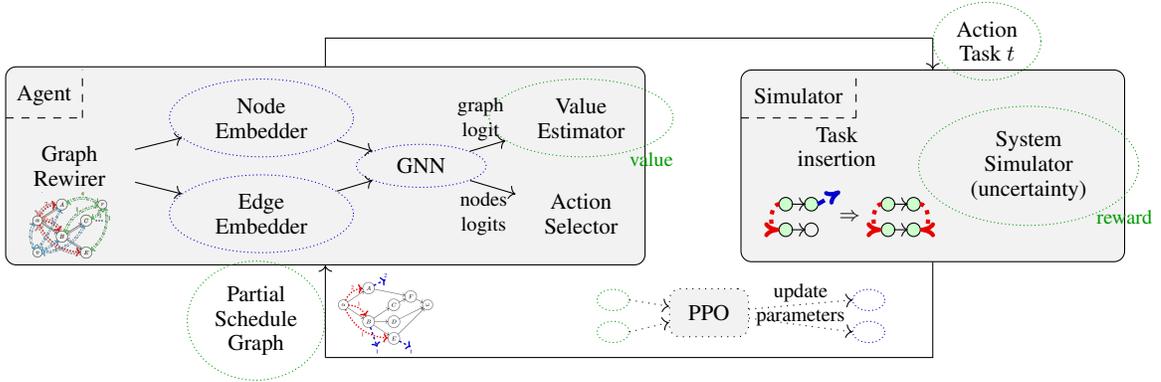
\begin{figure*}[!htbp]
  \centering

  \begin{tikzpicture}[scale=.85,every text node part/.style={align=center},every node/.style={scale=0.85}]
  	\tikzstyle{elem} = [draw=none,fill=none,text width=1.8cm]
  	
  	\begin{scope}
  		\node[draw,rounded corners, rectangle, minimum width=10cm, minimum height=3.1cm,fill=black!5]  (agent) at (4,0) {};
  		
  		\node[elem] (graphRewirer) at (0,0) {Graph\\Rewirer};
  		\node[elem,draw=blue!80!black,ellipse,densely dotted] (nodeEmbed) at (3,.75) {Node\\Embedder};
  		\node[elem,draw=blue!80!black,ellipse,densely dotted] (edgeEmbed) at (3,-.75) {Edge\\Embedder};
  		\node[elem,text width=1.2cm,draw=blue!80!black,ellipse,densely dotted] (gnn) at (5.5,0) {GNN};
  		\node[elem,draw=green!60!black,ellipse,densely dotted] (valEstim) at (8,.75) {Value\\Estimator};
            \node[green!60!black] at (9.1,0.1) {\footnotesize value};
  		\node[elem] (actSelector) at (8,-.75) {Action\\Selector};
  		
  		\node at (0,-.95) {\includegraphics[width=1.3cm]{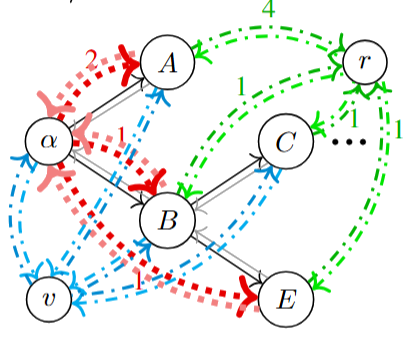}};
  		
  		\draw[->] (graphRewirer) -- (nodeEmbed);
  		\draw[->] (graphRewirer) -- (edgeEmbed);
  		\draw[->] (nodeEmbed) -- (gnn);
  		\draw[->] (edgeEmbed) -- (gnn);
  		\draw[->] (gnn) -- (valEstim) node [pos=.3, above] {\footnotesize graph\\ \footnotesize logit};
  		\draw[->] (gnn) -- (actSelector) node [pos=.3, below] {\footnotesize nodes\\ \footnotesize logits};;

  		\node at (-0.4,1.1) {Agent};
  		\draw[dashed] (-1,0.75) -- (.2,0.75);
  		\draw[dashed] (.2,0.75) -- (.2,1.5);
  	\end{scope}
  	
  	\begin{scope}[xshift=12cm]
  		\node[draw,rounded corners, rectangle, minimum width=6cm, minimum height=3cm, fill=black!5] (simu)  at (1.5,0) {};

  		\node[elem] (graphUpdate) at (0,0.3) {Task insertion};
  		\node[elem,text width=2.2cm,draw=green!60!black,ellipse, densely dotted] (sysSimu) at (3,0) {System Simulator (uncertainty)};

        \node[green!60!black] at (4.5,-.8) {\footnotesize reward};
  		
  		\begin{scope}[xshift=-.1cm]
  			\draw[] (-.3,-1) circle (3pt);
  			\draw[fill=green!20] (-.7,-.6) circle (3pt);
  			\draw[fill=green!20] (-.3,-.6) circle (3pt);
  			\draw[fill=green!20] (-.7,-1) circle (3pt);
  			\draw[->] (-.61,-1) -- (-.39,-1);
			\draw[->] (-.61,-.6) -- (-.39,-.6);
  			\path (-.8,-.6) edge[flows,-{>[length=2mm, width=2mm]},bend right=90] (-.8,-1);
                \path (-.2,-.6) edge[open_flows,-{>[length=2mm, width=2mm]}] (0.15,-.4);
  		\end{scope}
  	
  		\node at (0.2,-.8) {\footnotesize $\Rightarrow$};

		\begin{scope}[xshift=1.5cm]
			\draw[fill=green!20] (-.3,-1) circle (3pt);
  			\draw[fill=green!20] (-.7,-.6) circle (3pt);
  			\draw[fill=green!20] (-.3,-.6) circle (3pt);
  			\draw[fill=green!20] (-.7,-1) circle (3pt);
  			\draw[->] (-.61,-1) -- (-.39,-1);
			\draw[->] (-.61,-.6) -- (-.39,-.6);
  			\path (-.8,-.6) edge[flows,-{>[length=2mm, width=2mm]},bend right=90] (-.8,-1);
                \path (-.2,-.6) edge[flows,-{>[length=2mm, width=2mm]},bend left=90] (-.2,-1);
		\end{scope}

  		\node at (-0.6,1.1) {Simulator};
  		\draw[dashed] (-1.5,0.75) -- (.3,0.75);
  		\draw[dashed] (.3,0.75) -- (.3,1.5);
  	\end{scope}

        \coordinate (b) at (0,2);
  	
  	\draw (agent.north) -- (agent.north |- b) -- (simu.north |- b);
  	\draw[->] (simu.north |- b) -- (simu.north) node [pos=0.1, right, draw=green!60!black,ellipse, densely dotted] (action) {Action\\ Task $t$};

        \coordinate (a) at (0,-3);
  	
  	\draw (simu.south) -- (simu.south |- a) -- (agent.south |- a);
  	\draw[->] (agent.south |- a) -- (agent.south) node [pos=.4, left,draw=green!60!black,ellipse,densely dotted] {Partial\\Schedule\\Graph};
  	
  	\node[anchor=east] at (5.8,-2.3) {\includegraphics[width=1.5cm]{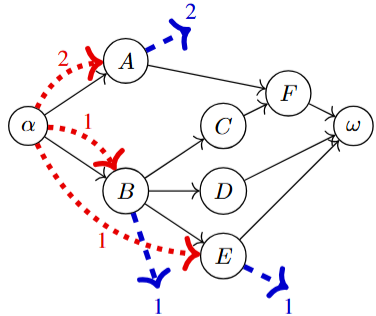}};
  	
  	\node[text centered, rounded corners, dotted, rectangle, draw, text width=1cm, minimum height=.75cm, fill=black!5] (ppo) at (10,-2.3)  {PPO};

        \node[draw=green!60!black,ellipse,densely dotted,minimum width=0.5cm] (input1) at (8.5,-2.1) {};

        \node[draw=green!60!black,ellipse,densely dotted,minimum width=0.5cm] (input2) at (8.5,-2.6) {};

        \draw[->, dotted] (input1.east) -- (ppo);
        \draw[->, dotted] (input2.east) -- (ppo);

        \node[draw=blue!80!black,ellipse,densely dotted,minimum width=0.5cm] (output1) at (12.5,-2.1) {};

        \node[draw=blue!80!black,ellipse,densely dotted,minimum width=0.5cm] (output2) at (12.5,-2.6) {};

        \draw[->, dotted] (ppo) -- (output1.west) node [pos=.5] {\footnotesize update \\  \footnotesize parameters};
        \draw[->, dotted] (ppo) -- (output2.west);
  	

  	
  	
  \end{tikzpicture}
  \caption{General Architecture - PPO's inputs and outputs are respectively surrounded by green and blue dotted ellipses.}
  \label{fig:archi}
\end{figure*}

       A general overview is shown in Figure~\ref{fig:archi}.

	\subsection{Graph Rewiring}
       The most used paradigm for GNN is the ``message-passing'' paradigm. Within this algorithm family, messages are computed at node level, and then propagated along edges. Destination nodes then aggregate incoming neighbor messages and update their own values.  Here, the graph structure serves as a computational lattice, strongly constraining messages pathways. This leads to two well-known issues: over-squashing (occurring when a single edge is part of multiple paths), and over-smoothing (arising when a node must aggregate/filter data from an excessive number of paths) - see \cite{akansha2025oversquashinggraphneuralnetworks} for a recent review and analysis. To mitigate these problems, a common approach involves performing \emph{graph rewiring}, which entails modifying the graph (typically by adding edges) in order for the lattice to be able to convey data (through messages) to necessary locations for the terminal task.
	
	\paragraph{MDP graph and bidirectional message-passing}
        A critical issue in our MDP graph is observed: the directed edges, representing precedences (from  task constraints or resource consumption order choices), flow from past to future. This lattice structure poses a problem, as it prevents messages from future conflict patterns from propagating back to inform present candidate choices - a crucial functionality for our purpose. To address this, we augment our MDP graph by incorporating the inverse edges of both the precedence graph and the flow graph (with the same attributes). Formally, for every arc $(t_1,t_2) \in \precedences$, we add its counterpart $(t_2, t_1)$, and for every flow arc $(r,k,t_1,t_2)$ we add the reciprocal edge $(r,k,t_2,t_1)$. To enable the GNN to differentiate between the original and newly added edges, we also assign a distinct type to each edge category (as detailed below). 
	
	\paragraph{Edge Dropping}
       Upon task scheduling, their completion times are computed, making past flow graphs no longer useful for future scheduling decisions. To enhance the agent's efficiency, we prune past edges, \textit{i.e.} all edges that lead to scheduled nodes (both precedence and flow graph edges). 
	
	\paragraph{Resources as Nodes}
       To schedule tasks in a way facilitating efficient future decisions, the agent must have a representation of future constraints and conflicts (to learn useful patterns). Relying solely on flow graphs for past or present tasks (\textit{i.e.} those currently being scheduled) may not provide sufficient information. To explicitly highlight potential upcoming conflicts, we introduce resource nodes into the graph, accompanied by edges representing consumptions. Formally, for each $r \in \mc{R}$, we add a node $n_r$, and for each consumption $\mi{cons}_{t,r}$
       of resource $r$ by a task $t$, we add edges $(n_r,t)$ and $(t,n_r)$ with attribute $\mi{cons}_{t,r}$. For performance considerations, and without loss of generality, such edges are not added for tasks already scheduled. An alternative approach could involve direct connections between task nodes sharing the same resource, resulting in  a clique per resource. However, the associated number of edges would quickly make the graph untractable.

	\paragraph{Graph Pooling as a Virtual Node}
       To compute a complete graph pooling, conventional methods typically involve aggregating all node information using simple operators (like max, mean) or more complex operations that account for the structure. In our approach, we take a middle ground approach by adding a virtual node $v$ connected to both task nodes $t_i$ and resources nodes $n_r$ through directed edges. Formally, for each $t_i \in \tasks$, an edge $(t_i, v)$ is added; for each resource node $n_r , r\in \mc{R}$, an edge $(n_r, v)$ is also added. These edges will have different type, see below. This enables the GNN to learn its own pooling operator while computing messages and values for this node concurrently with other nodes.

	\paragraph{Typing Everything}
       The resulting rewired graph comprises a diverse set of nodes and edges. Specifically, on the node side, we have: task nodes, resource nodes and a global virtual pool node. On the edge side, the following types are present: task precedence edges; reverse task precedence edges; flow graph edges; reverse flow edges; links from tasks to resource nodes (for future potential conflicts); links from resources nodes to tasks (the converse of the previous); links from tasks to the virtual pool node; links from resource nodes to the virtual pool node; self loops for tasks; self loops for resource nodes and self loops for the pool node.

To enable the GNN to convey useful data across semantically distinct edges, we assign an explicit type to each edge. Moreover, we require a message-passing mechanism that is able to pass different messages based on both the node and edge types.

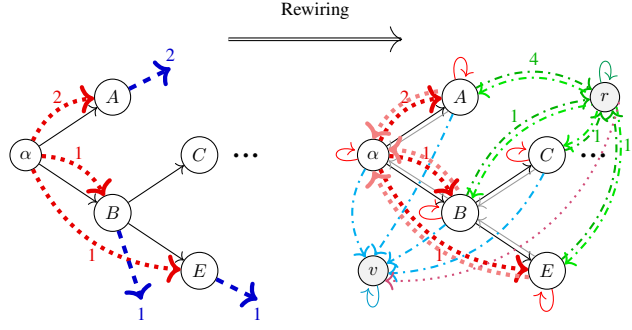
\begin{figure}
        \centering
        \begin{tikzpicture}[scale=0.77, every node/.style={scale=0.7}]
            \begin{scope}
				\node[circle,draw] (alpha) at (0,0) {$\alpha$};
				\node[circle,draw] (a) at (1.5,1) {$A$};
				\node[circle,draw] (b) at (1.5,-1) {$B$};
				\node[circle,draw] (c) at (3,0) {$C$};
				\node[circle,draw] (e) at (3,-2) {$E$};
                    \node at (3.8,0) {\huge...};
				
				\draw[->] (alpha) -- (a);
				\draw[->] (alpha) -- (b);
				
				\draw[->] (b) -- (c);
				
				\draw[->] (b) -- (e);

                    \path[->, flows] (alpha) edge[bend left] node[above] {2} (a);
                    \path[->, flows] (alpha) edge[bend left] node[above] {1} (b);
                    \path[->, flows] (alpha) edge[bend right] node[below] {1} (e);
                    \path[->, open_flows] (b) edge[] node[below,pos=1] {1} (2,-2.5);
                    \path[->, open_flows] (e) edge[] node[pos=1,below] {1} (4,-2.5);
                    \path[->, open_flows] (a) edge[] node[above,pos=1] {2} (2.5,1.5);
                \end{scope}

                \begin{scope}[xshift=6cm]
				\node[circle,draw] (alpha) at (0,0) {$\alpha$};
                    \path (alpha) edge[self_loop_task,->,loop left] (alpha);
				\node[circle,draw] (a) at (1.5,1) {$A$};
                \path (a) edge[self_loop_task,->,loop above] (a);
				\node[circle,draw] (b) at (1.5,-1) {$B$};
                \path (b) edge[self_loop_task,->,loop left] (b);
				\node[circle,draw] (c) at (3,0) {$C$};
                \path (c) edge[self_loop_task,->,loop left] (c);
				\node[circle,draw] (e) at (3,-2) {$E$};
                \path (e) edge[self_loop_task,->,loop below] (e);
                    \node at (3.8,0) {\huge...};

                    \node[circle,draw,fill=gray!10]  (r) at (4,1) {$r$};

                    \path (r) edge[self_loop_resource,->,loop above] (r);
                    
                    \node[circle,draw,fill=gray!10]  (v) at (0,-2) {$v$};

                    \path (v) edge[self_loop_pool,->,loop below] (v);
				
				\draw[->] (alpha) -- (a);
				\draw[->] (alpha) -- (b);
				
				\draw[->] (b) -- (c);
				
				\draw[->] (b) -- (e);

                    \draw[<-,back_prec] (alpha.20) -- (a.230);
				\draw[<-,back_prec] (alpha.340) -- (b.130);
				
				\draw[<-,back_prec] (b.20) -- (c.230);
				
				\draw[<-,back_prec] (b.340) -- (e.130);

                    \path[->, flows] (alpha) edge[bend left] node[above] {2} (a);
                    \path[->, flows] (alpha) edge[bend left] node[above] {1} (b);
                    \path[->, flows] (alpha) edge[bend right] node[below] {1} (e);

                    \path[<-, back_flows] (alpha.90) edge[bend left] (a.160);
                    \path[<-, back_flows] (alpha.20) edge[bend left] (b.90);
                    \path[<-, back_flows] (alpha.270) edge[bend right](e.200);

                    \path[->,resource] (a) edge[bend left] node[above] {4} (r);
                    \path[<-,back_resource] (a.10) edge[bend left] (r.170);

                    \path[->,resource] (b) edge[bend left] node[above] {1} (r);
                    \path[<-,back_resource] (b.50) edge[bend left] (r.210);

                    \path[->,resource] (c) edge[bend right] node[below,pos=.8] {1} (r);
                    \path[<-,back_resource] (c.30) edge[bend right] (r.240);

                    \path[->,resource] (e) edge[bend right] node[right,pos=.8] {1} (r.300);
                    \path[<-,back_resource] (e.20) edge[bend right] (r.320);

                    \path[->,pooling] (alpha) edge[bend right] (v);

                    \path[->,pooling] (a) edge[bend left=5] (v.85);

                    \path[->,pooling] (b) edge[bend left=5] (v);

                    \path[->,pooling] (c.260) edge[bend left] (v.350);

                    \path[->,purple!70,dotted,thick] (r.340) edge[bend left=40] (v.320);
                    
                \end{scope}

                \draw[double,->] (3.5,2) -- (6.5,2);
                \node at (5,2.5) {Rewiring};
                
			\end{tikzpicture}
        \caption{Rewiring of the graph presented in Figure~\ref{fig:AON-flow_insertion}. Nodes $r$ and $v$ represent the resource node for $r$ and the pooling node. Back arcs of flows are in dotted light red, arcs to and from resource $r$ are in dashdotted green, and arcs to the pooling node are in dashdotted cyan for activity nodes, dotted purple for resources nodes.} 
        \label{fig:rewiring}
    \end{figure}

\begin{example}
    Figure~\ref{fig:rewiring} illustrates the rewiring of the graph in the state $s_i$. Nodes associated with the unique resource $r$ and the graph pooling $v$ are added, and so are all backwards edges, along with their respective types. 
\end{example}

	\subsection{Handling Uncertainty}
       As outlined in Section \ref{sec:repr_unc}, duration uncertainty is captured by maintaining three distinct flow graphs, respectively associated with the  minimum, maximum and mode (or average) durations. These graphs are provided to the agent as an observation. On the simulator's end, an additional flow graph is generated based on one possible outcome of the policy, where task durations are randomly drawn. This results in an observed total duration for each trial, which serves as a reward signal for the RL algorithm. Notably, this reward signal is only provided at the end of the trial (\textit{i.e.} for a complete schedule), as a terminal reward.

	\subsection{Implementation Details}
       This section give some details on precise implementation. 

	\paragraph{Task Nodes Attributes}
       The node have the following attributes as input for the GNN:
       \begin{itemize}
         \item ``affected'': initially set to 0, and set to 1 for already scheduled tasks;
         \item ``selectable'': set to 1 when all precedences are ``affected'', and reset to 0 when this tasks is scheduled;
           \item ``past'': is set to one when a tasks is affected and does not have any more open flow arcs in any flow graph; 
         \item ``type'': set to -1 for source node, 1 for sink nodes and 0 for all others;
         \item ``normalized durations'': the durations (min, max and mode) of the tasks divided by the maximum duration of all tasks;
         \item ``normalized task completion time'': the computed tasks completion times (min max and mode)  given the current partial schedule (divided by the tasks maximum duration);
       \end{itemize}
       All these attributes are embedded using a MLP before being provided to the GNN. 

       \paragraph{Resource Nodes Attributes}
       In contrast, resource nodes do not have any attribute. In fact, the only relevant attribute would be the resource capacity, which is normalized internally by the simulator, meaning that all resources have a capacity equal to 1. 
       
       \paragraph{Edge Attributes}
	For most edges, their type constitutes their only attribute. However, there are exceptions for resource levels regarding (reverse) flow graph edges and links to and from resource nodes. In these cases, the edge type is encoded using an embedding, while the resource levels are encoded with a simple linear layer. Both are summed to obtain the final embedding.

       \paragraph{Message Passing with Attention}
       One distinctive feature of our graph is its inherent heterogeneity, comprising multiple edge types, and three node types. To compute semantically different quantities, the GNN must be able to pass different messages across various edge types, coming from different node types. This necessitates a message passing mechanism that can build very different messages. We employ Gatv2 with edges attributes \cite{gatv2}. This mechanism learns an attention weight applied to the embeddings of both source and destination nodes as well as the edge itself, meaning that all three types and attributes can be used for the message construction. Formally, a node $x_i$ is updated as follows: $x_i'= \sum_{j \in \mc{N}(i) \cup \{i\}} \alpha_{i,j}\theta_tx_j$, where $x_j$ are the neighbors (source nodes) $\theta_t$ is a learned linear projection, and attention coefficients $\alpha_{i,j}$ are computed as : $\alpha_{i,j} = \frac{\text{exp}(a^\intercal \text{LeakyReLU}(\theta_sx_i + \theta_tx_j + \theta_e e_{i,j}))}{\sum_{k \in \mc{N}(i) \cup \{i\}}\text{exp}(a^\intercal \text{LeakyReLU}(\theta_sx_i + \theta_tx_k + \theta_e e_{i,k}))}$. In this definition, $a$ is a learned coefficient vector, $\theta_e$ and $\theta_s$ are learned linear projections.

	\section{Experiments}
	\label{sec:expes}
	\subsection{Wheatley Setup Details}
    We provide details for Wheatley, denoted $\texttt{W}$, which is publicly available \cite{wheatley}. 
    \paragraph{GNN} The GNN itself is implemented using Pytorch Geometric \cite{pyg}. It is made up of 8 GATv2 layers \cite{gatv2} with residual connections. After the GNN part, all layers node logits are concatenated and passed through a two-layers MLP, which gives the actor networks. For the critic network, all layers virtual node logits are concatenated and given to a two-layers MLP, that outputs a single value. We employ the Gelu \cite{gelu} activation function everywhere. The node values at first layers are embedded using a simple MLP with 3 layers, the edge values are embedded by summing a simple embedding of their type to an embedding of their attributes obtained by passing these attributes through a MLP. While the values of the nodes are computed by the GNN, the edge embedding is not updated by the GNN; we thus have to pass embedded values to the edges at every layer. We found it better to have an embedder per layer instead of a single embedder for every layer. 
    \paragraph{RL} we implemented PPO \cite{PPO} starting from CleanRL \cite{cleanRL} implementation. All rewards are normalized so that it is the task completion time of last task divided by the number of jobs, giving cost values a small magnitude, in our case between $-2$ and $-1$.  Node features contain normalized durations and task completion times, obtained by dividing the durations and task completion times by the maximum duration of all tasks. Such a normalization allows to remove the gradient clipping within PPO. Furthermore, by using a batch size of 256 and gathering $500 000$ steps of (action,observation) pairs at every iteration, we were able to remove the entropy coefficient from PPO loss, which is generally used to prevent the agent from falling too early into local minima. 
    \paragraph{Training} Every $500 000$ steps data set is processed three times, and it never reaches the target KL divergence bound, which is sometimes used to trigger early update stopping. Finally, we employ Radam Optimizer \cite{radam} with a decoupled weight decay of 0.01. 
    Note that Wheatley is trained once for all instances during one day on one machine, and the evaluation inference time for largest instances lasts $20$ seconds.

	
	\subsection{Benchmarks}
    We use benchmarks provided in the PSPLib~\cite{kolisch1997psplib} that are composed of instances with 30, 60, 90 and 120 tasks. Following~\cite{VerhaegheCPQ24}, we divide each set of instances of given size into three subsets: one training set, one unseen set (\textit{usn}) and one unknown set (ukn). The usn subset corresponds to instances that have not being used for training but similar to the training one (same configuration but different seed). The ukn subset contains instances with a structure that has not been used during training. This results in $1,472$ training instances ($3 \times 344$ for 30, 60 and 90 tasks + $440$ for 120 tasks), $368$ usn instance ($3 \times 86$ for 30, 60 and 90 tasks + $110$ for 120 tasks) and $200$ ukn instances ($4 \times 50$). 

    To evaluate the approaches in the uncertain case, we have generated $100$ duration scenarios for each usn and ukn instance, meaning that each approach is tested on $36,800$ usn instances and $20,000$ ukn instances. 
    
        \subsection{Baselines Experimental Setup}
        \paragraph{Deterministic Case}
        Although the primary objective of our work is to address uncertainty, we also compare our approach against the best solution known in the deterministic counterpart of each instance. 

        \paragraph{Non-deterministic Case}
        We compare our approach with deterministic-based solving approaches that we have adapted in order to address uncertainty. For each of the following baselines, the idea is the same: generate a priority rule and for each duration scenario of each instance, apply this priority rule within a SSGS using the real tasks duration. 
        We use the following baselines: \\
        \textbf{\emph{Priority Dispatch Rules}} When needed in the ranking process, we use the mode duration of tasks as tasks durations. The rules we have considered are: 
            \begin{itemize}
                \item SPT /LPT (Shortest/Longest Processing Time): tasks with the shortest/longest duration first;
                \item MIS (Most Immediate Successors): tasks with the greatest number of direct successors first;
                \item GRPW (Greatest Rank Positional Weight): tasks for the which the sum of the task duration and the successors duration first;
            \end{itemize}
         \textbf{\emph{CP-50}}: this approach relies on a Constraint Programming (CP) encoding of the problem. Such an approach decides of the start time of each task (encoded through interval variables), while ensuring precedence constraints and resources capacities constraints (encoded with \emph{Cumulative} constraints). In this baseline, we generate a set of 50 duration scenarios and enforce the use of the same priority list in each scenario so that the average makespan mean value is minimized when applying SSGS. There is no trivial way to encode generation schemes through CP. Thus, we first model the priority list with a integer variable $\mi{pos}_t$ for each task $t$ representing its position in the list. Then, we enforce that the position variables are consistent with precedence constraints. Moreover, if two tasks $t_1$ and $t_2$ consume at least one common resource and $\mi{pos}_{t_1} < \mi{pos}_{t_2}$, then $t_1$ starts before $t_2$ in the schedule ($\mi{startBeforeStart}$ constraint). Finally, the solver is tuned such that the variable selection heuristics first chooses $\mi{pos}_t$ variables, which improves greatly the solving performance; \\
        \textbf{\emph{CP-50-ws}}: on larger instances, the CP solver can struggle to find an initial solution. Therefore, we propose the \emph{CP-50-ws} baseline that first computes a schedule using a PDR (we use here SPT), maximum tasks durations and no overlaps between tasks. This solution is always feasible and we use it to warm-start the \emph{CP-50} approach. 
        
        CP-based approaches were implemented with CP Optimizer \cite{laborie2018ibm}, version 22.1.1 used through the DOcplex API, run on Intel® Xeon® CPU E5-2660v3 2.60-3.30 GHz with 62 GB of RAM with a time-out of 60 seconds.
    
	\subsection{Results}
        \paragraph{Deterministic Case}
        Table~\ref{tab:results_deterministic} presents results associated with Wheatley in comparison with the best solution found. Each cell in the table represents the average gap value for all instances of the dataset, where the gap is equal to $100\cdot\frac{\mi{sol}-\mi{best}}{\mi{sol}}$. Experimental results show that the gap less than $5\%$ for smaller datasets (30, 60 and 90) but increases greatly in the case of 120 jobs datasets. 
    
	\begin{table}[t]
		\centering
		\small
		\setlength{\tabcolsep}{5pt}
		\begin{tabular}{c|cccc|c}
			evaluation set & 30 & 60 & 90 & 120 & average\\
			\hline
			     usn & 4.42 & 4.36 & 4.62 & 10.39 & 6.24 \\
			     ukn & 4.19 & 4.61 & 4.28 & 12.48 & 6.39\\
		\end{tabular}
        \vspace{5mm}
		\caption{Results on deterministic problems, average GAP(\%) compared to best solution known} \label{tab:results_deterministic}
	\end{table}

    \paragraph{Uncertain Case}
    Table~\ref{tab:uncertain} presents the average makespan (\textit{mks} column) obtained for all scenarios for each benchmarks subset. The associated gap with respect to the best solution found is indicated in the gap column, and the column \emph{cov} represents the percentage of scenarios for which the approach has found a solution. When absent, the default coverage value is 100\%. 
    Obtained results show that our approach obtains the best solution in most benchmarks subsets. More precisely, PDRs are not able to provide efficient priority rules, as they are quite far from the best solution found. When comparing with CP-based approaches, the results depend on the size of the datasets. In fact, for the smallest datasets (usn 30 and ukn 30), the gaps between \texttt{DRL} and CP-50 are quite close. This is expected as the 50 scenarios can be handled by the exact approach. However, with larger datasets, CP-based approaches struggle to find solutions. Note that CP-50 and CP-50-ws achieve higher average scores for the instances they solve, albeit with a significantly lower number of solved instances, rendering them impractical for real-world applications.
    For datasets with 90 or 120 jobs, the gap increases greatly for CP-50, and it even reaches time out for the 120 jobs ukn datasets. The CP-50-ws starts from an initial solution and is able to find more solutions than CP-50. However, the gap with Wheatley is still greater than $12\%$ for the ukn 120. These results demonstrate that Wheatley can effectively handle not only datasets similar to those used in training, but also novel datasets never encountered before. 
    
    \begin{table*}[ht]
        \centering
        \tiny
        \setlength{\tabcolsep}{4pt}

        \begin{tabular}{cccccccccccccccccc}
        \toprule
        dataset & \multicolumn{3}{c}{\texttt{W}} & \multicolumn{3}{c}{CP-50} & \multicolumn{3}{c}{CP-50-ws} & \multicolumn{2}{c}{SPT} & \multicolumn{2}{c}{LPT} & \multicolumn{2}{c}{MIS} & \multicolumn{2}{c}{GRPW} \\
        \cmidrule(lr){2-4}\cmidrule(lr){5-7}\cmidrule(lr){8-10}\cmidrule(lr){11-12}\cmidrule(lr){13-14}\cmidrule(lr){15-16}\cmidrule(lr){17-18}
        & mks & gap & cov & mks & gap & cov & mks & gap & cov& mks & gap & mks & gap & mks & gap & mks & gap \\
        \midrule 
        usn 30 & \textbf{64.01} & \textbf{1.3} & 100 &64.14 & 2.6 & 100 & 68.85& 7.7& 100 & 79.07 &24.8& 79.19 &24.9 & 72.49 &14.2&  72.26 &21.8 \\
        ukn 30 & \textbf{71.39} & \textbf{0.7} & 100& 73.14  &3.3 &100  & 77.23 &8.0& 100  &  82.65 &16.1 & 82.75 &15.3 & 78.77 &10.0 & 79.49 &11.5 \\
        usn 60 & \textbf{87.32} & \textbf{0.1} &100 & 101.36 &15.0 & 100 & 112.28 & 26.8 & 100 & 113.78 &28.8 & 114.11 &29.7 & 102.34 &16.7 & 111.36 &26.4 \\
        ukn 60 & \textbf{96.72} & \textbf{0.0} & 100& 108.83 & 11.8 & 100  & 116.62 & 18.9& 100  & 116.62 &18.9 & 113.74 &15.8 & 107.09 &10.1 &  112.75 &15.3 \\
        usn 90 & \textbf{105.08} &\textbf{1.2} & 100 &124.54 &19.1& 98.8 & 140.84 & 31.8 & 100 & 140.84 &31.8 & 140.96 &32.1 & 124.35 &17.0 &  136.23 &27.5 \\
        ukn 90 & 117.21 & 0.1 & \textbf{100}& \textbf{111.47}& 9.4 &90 & 144.40 &21.1& 100 & 144.40 &21.1&  139.61 &16.9 & 133.21 &11.9 & 138.86 &11.12 \\
        usn 120 & 141.47 & \textbf{0.0} & \textbf{100} &\textbf{62.17}&  15.3 &40 & 160.33 & 40.8 &74.6 &214.34 &51.5 &212.73 &50.5& 181.90 &28.0 & 206.68 &46.3 \\
        ukn 120 & 187.99 & \textbf{0.0} & \textbf{100}& - & - & 0 & \textbf{96.30}& 12.2 &34  & 259.45 &38.5 & 250.84 &32.4 & 230.95 &22.7 & 245.4 &30.7 \\
        \bottomrule
        \end{tabular}
        \vspace{5mm}
        \caption{Average makespans (mks), gaps (gap, \%) compared to best solution found and percentage of problem solved  before timeout (cov, \%). (SPT, LPT, MIS and GRPW coverage is always 100\%) } \label{tab:uncertain}
    \end{table*}

	\section{Conclusion and Future Work} \label{sec:conclusion}
    In this work, we introduce a novel approach for solving RCPSP with uncertain task duration, leveraging GNN and DRL. Our framework, Wheatley \cite{wheatley}, relies on a sophisticated encoding of flow and resource information within the GNN. Notably, it outperforms both existing Priority Dispatching Rules (PDRs) approaches and deterministic-based strategies. 
    
    Several extensions are foreseen for this work. Preliminary experiments conducted on non-public industrial datasets have demonstrated promising scalability on very large datasets but these findings require validation through a more comprehensive set of benchmarks. Additionally, considering multimodal distributions would necessitate further studies. 
    Finally, updating the approach to accommodate alternative criteria beyond makespan and some RCPSP variants constraints is straightforward but would require additional experiments.  

    \bibliographystyle{plain}
    \bibliography{biblio}

\end{document}